\documentclass[letterpaper]{article} 
\usepackage{aaai2026}  
\usepackage{times}  
\usepackage{helvet}  
\usepackage{courier}  
\usepackage[hyphens]{url}  
\usepackage{graphicx} 
\usepackage{natbib}  
\usepackage{caption} 
\usepackage{algorithm}
\usepackage{algorithmic}
\usepackage{amssymb}
\usepackage{algorithm}
\usepackage{algorithmic}
\usepackage{booktabs} 
\usepackage{multirow} 
\usepackage{graphicx}
\usepackage{array}
\newcolumntype{L}[1]{>{\raggedright\let\newline\\\arraybackslash\hspace{0pt}}m{#1}}
\newcolumntype{C}[1]{>{\centering\let\newline\\\arraybackslash\hspace{0pt}}m{#1}}
\newcolumntype{R}[1]{>{\raggedleft\let\newline\\\arraybackslash\hspace{0pt}}m{#1}}
\usepackage{amsmath}
\usepackage{newfloat}
\usepackage{listings}
\DeclareCaptionStyle{ruled}{labelfont=normalfont,labelsep=colon,strut=off} 
\floatstyle{ruled}
\newfloat{listing}{tb}{lst}{}
\floatname{listing}{Listing}
\title{Selective Weak-to-Strong Generalization}
\author{
    Hao Lang,
    Fei Huang,
    Yongbin Li \thanks{Corresponding author.}
}
\affiliations{
    Tongyi Lab\\
    \{hao.lang, f.huang, shuide.lyb\}@alibaba-inc.com
}

\usepackage{bibentry}

\begin{document}

\maketitle

\begin{abstract}
Future superhuman models will surpass the ability of humans and humans will only be able to \textit{weakly} supervise superhuman models.
To alleviate the issue of lacking high-quality data for model alignment, some works on weak-to-strong generalization (W2SG) finetune a strong pretrained model with a weak supervisor so that it can generalize beyond weak supervision.
However, the invariable use of weak supervision in existing methods exposes issues in robustness, with a proportion of weak labels proving harmful to models.
In this paper, we propose a selective W2SG framework to avoid using weak supervision when unnecessary.
We train a binary classifier P(IK) to identify questions that a strong model can answer and use its self-generated labels for alignment.
We further refine weak labels with a graph smoothing method.
Extensive experiments on three benchmarks show that our method consistently outperforms competitive baselines.
Further analyses show that P(IK) can generalize across tasks and difficulties, which indicates selective W2SG can help superalignment.
\end{abstract}

\section{Introduction}

Most existing AI alignment methods rely on the availability of human labelled data, such as human demonstrations for SFT~\citep{wei2021finetuned}, or human preferences for RLHF~\citep{christiano2017deep,ouyang2022training}.
These methods have been used to build the most capable AI systems currently deployed~\citep{OpenAI2023Gpt-4,Anthropic2023Introducing}.

However, future superhuman models will surpass the ability of humans in certain areas and behave in complex ways that humans cannot reliably evaluate~\citep{CAIS2023Statement}.
For example, if a superhuman model generates a billion lines of complicated code, humans will not be able to provide reliable supervision for superalignment (aligning superhuman models).
In that case, the expected deficiency of human evaluation will limit the effectiveness of most alignment methods~\citep{casper2023open}.

\begin{figure}[t]
\centering
\includegraphics[width = 0.93\linewidth]{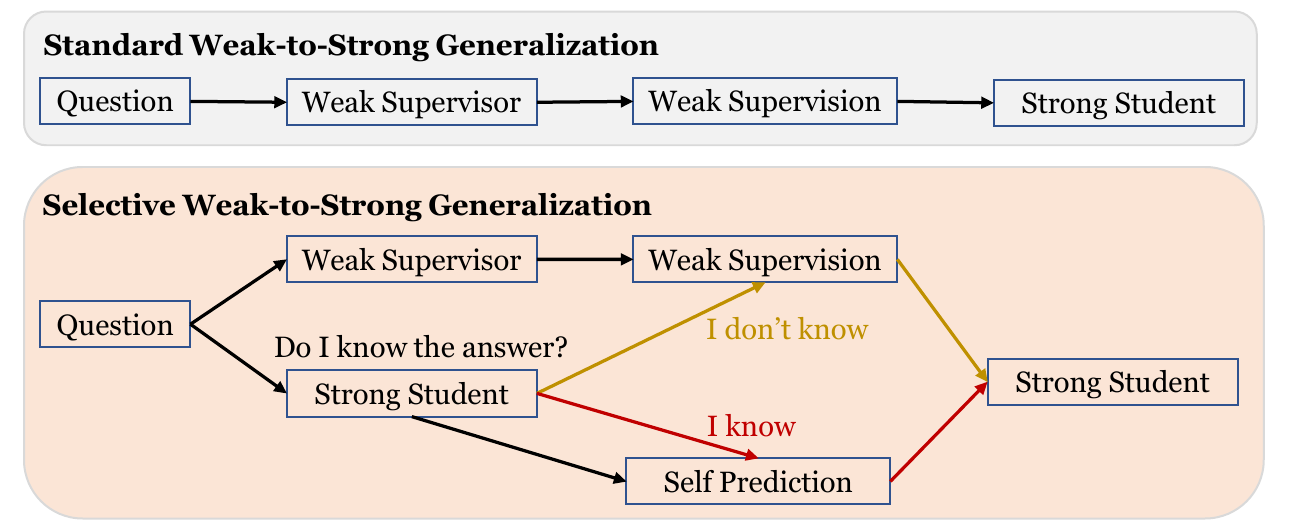}
\caption{\textbf{Comparison of standard weak-to-strong generalization  (W2SG) (top) and selective W2SG (bottom)}. Different from (top), which always uses weak supervision to train the strong model, we instead use the prediction of the strong model to train itself when the strong model knows the correct answer.}
\label{fig:diagram}
\end{figure}

In response to the challenge of superalignment, recent work on weak-to-strong generalization (W2SG) has started to explore the potential of finetuning a strong pretrained model with a weak supervisor~\citep{burns2023weak}.
In this study, a strong pretrained model is assumed to already have good representations of the alignment-relevant tasks, and a weak supervisor is expected to elicit what the strong model already knows.
Empirical results show that the strong model can generalize beyond the weak supervision and even outperforms its weak supervisor on specific tasks given flawed training labels (Figure~\ref{fig:diagram} top).

Despite their encouraging performance, existing W2SG approaches largely ignore to address a critical question:
\begin{itemize}
    \item \textit{Should we always use weak supervision to train the strong model?}
\end{itemize}

Our findings suggest that the answer is predominantly negative.
First, in various tasks, the weak supervisor can only give incomplete or flawed training labels with noise.
The desired generalization should be able to disagree with the weak supervision when the weak supervision is wrong~\citep{burns2023weak}.
Second, existing W2SG approaches are still far from recovering the full capabilities of strong models.
It suggests that alignment methods may scale poorly to superhuman models without additional work.

In this paper, we challenge the assumption of always using weak supervision by proposing a novel superalignment framework underpinned by a \textit{selective weak-to-strong generalization} mechanism: the system proactively abstains from training a strong model with potentially detrimental weak labels (Figure~\ref{fig:diagram} bottom).
At the core of our framework is a strong model trained to estimate P(IK), the probability that it knows the answer to a question, on a given distribution (see Table~\ref{tab:prompt2} for an illustration)~\citep{kadavath2022language}.
The framework confidently uses labels generated by the strong model for alignment, when the strong model already knows the correct answers with high P(IK) scores.

Specifically, we train a strong model with a binary classification head to predict P(IK), which elicits the capability of evaluating its own knowledge state.
We leverage additional existing datasets, generate strong model predictions, and obtain labels for classification by contrasting predictions with ground truth labels. 
Then, we jointly train a strong model to predict P(IK) and align with the optional weak supervision.

Furthermore, we refine weak supervision with labels generated by the strong model via a graph smoothing method~\citep{luo2018smooth,li2021comatch}.
Concretely, we divide questions into high P(IK) score IK questions with self-generated labels and low P(IK) score IDK questions with weak labels, respectively.
We construct a graph comprised of IK and IDK questions, and produce a graph-smoothed label for each IDK question by aggregating nearby nodes on the graph, based on the smoothness assumption~\citep{van2020survey}.

We perform comprehensive evaluations on three OpenAI weak-to-strong NLP benchmarks, which convert existing tasks to binary classification problems to make empirical progress in superalignment~\citep{burns2023weak}.
Results show that our method achieves strong performance, outperforming methods that always use weak supervision.

Finally, we provide comprehensive analyses on the generalization ability of the P(IK) classifier.
We show that P(IK) can generalize across tasks and difficulties, e.g., from science QA to reading comprehension, commonsense reasoning, and math.
These analyses indicate that selective W2SG can help align future superhuman models.


Our major contributions are summarized:

\textbf{1.} We challenge the assumption of always using weak supervision for W2SG and propose a selective W2SG framework for superalignment.

\textbf{2.} We train a binary classifier P(IK) to identify whether using weak labels is necessary. We also refine weak labels with a graph smoothing method. 

\textbf{3.} Extensive evaluations on three OpenAI weak-to-strong NLP benchmarks show that our approach outperforms competitive baselines. Further analyses indicate that the generalization ability of the P(IK) classifier is promising.

\begin{table*}[t]
    \centering
    \small
    \begin{tabular}{L{380pt} |  C{40pt} }
    \toprule
    \textbf{Problem} & \textbf{P(IK)} \\
    \midrule
    Q: What disease occurs when the cell cycle is no longer regulated? A: cancer R: & 99.6\% \\
    \midrule
    Q: Digestive enzymes are released, or secreted, by the organs of which body system? A: digestive system & 99.6\% \\
    \midrule
    Q: Specific antigens on the surface of red blood cells determine what, which is important in cases of transfusion? A: blood type R: & 99.6\% \\
    \midrule
    Q: What forms when nitrogen and sulfur oxides in air dissolve in rain? A: acid snow R: & 13.4\%\\
    \midrule
    Q: Movements in the mantle cause the plates to move over time in a process called what? A: continental shift R: & 13.2\%\\
    \midrule
    Q: What are the little sacs at the end of the bronchioles called? A: respiratory sacs R: & 12.7\%\\
    \bottomrule
    \end{tabular}
    \caption{\textbf{Examples of P(IK) scores from a 14B model}. Harder problems have lower P(IK) scores. The binary classification problem is converted from the SciQ dataset. We also append the custom prompt \textit{[There is a science knowledge question, followed by an answer. Respond with 1 if the answer is correct, and with 0 otherwise.]} to the problem. P(IK) refers to the probability that I Know the answer.}
    \label{tab:prompt2}
\end{table*}
\section{Related Work}

\textbf{AI alignment.} AI alignment aims to steer already-capable models to behave in line with human values and intentions~\citep{leike2018scalable,ji2023ai}.
Existing alignment methods finetune pretrained models using imitation learning on human demonstrations~\citep{bain1995framework,atkeson1997robot,wei2021finetuned,chung2024scaling}, reinforcement learning from human feedback (RLHF)~\cite{christiano2017deep,stiennon2020learning,ouyang2022training,bai2022training}, or direct alignment algorithms like direct preference optimization (DPO)~\citep{rafailov2024direct,rafailov2024scaling}.
Both imitation learning and preference learning rely on high-quality supervision, a demand that becomes increasingly challenging as models become more capable than humans~\citep{amodei2016concrete}.

\paragraph{Weak-to-strong generalization.} Weak-to-strong generalization techniques seek to finetune a strong pretrained model to generalize well from weak supervision~\citep{burns2023weak}.
This is typically pursued by assuming strong pretrained models should already have good representations of the alignment-relevant tasks, and thus we simply need a weak supervisor to elicit what the strong model already knows.
Recently, ~\citet{charikar2024quantifying,mulgund2025relating,xu2025emergence} theoretically quantify the gain in W2SG with misfit error.

A core question in this area is finding an effective policy to improve weak supervision.
~\citet{burns2023weak} propose an auxiliary confidence loss to reinforce the strong model’s confidence in its own predictions.
~\citet{guo2024vision} propose an adaptively adjustable loss function for vision superalignment.
~\citet{yao2025revisiting, yao2025weak} propose reverse KL and f-divergence losses to improve noise tolerance of the strong model.
~\citet{lyu2024macpo,ye2025iterative,lang2025debate} propose iterative updating methods, and ~\citet{liu2024co,cui2024bayesian,agrawal2024ensemw2s} propose ensemble methods.
In comparison, this paper highlights the importance of understanding whether a strong model knows the correct answer~\citep{kadavath2022language} when improving weak supervision.

\paragraph{Semi-supervised learning.} Our work is also related to semi-supervised learning (SSL) since they share the same \textit{smoothness assumption}, i.e., samples close
to each other are likely to receive similar labels~\citep{luo2018smooth,van2020survey}. The auxiliary confidence loss is also related to a key technique in SSL~\citep{burns2023weak}.

\section{Preliminaries}

We start by reviewing the weak-to-strong generalization pipeline in~\cite{burns2023weak}, which has also been adopted in subsequent work~\cite{charikar2024quantifying,guo2024vision}.
It usually consists of three steps:

\paragraph{1. Create the weak supervisor.} The first step is to create the weak supervisor by finetuning a small pretrained model on ground truth labels.
The performance of the weak supervisor is called the \textit{weak performance}.

\paragraph{2. Train a strong student model.} The second step is to train a strong student model by finetuning a large pretrained model on weak labels generated by the weak supervisor.
Its performance is called the \textit{weak-to-strong performance}.

\paragraph{3. Train a strong ceiling model.} The third step is to train a strong ceiling model by finetuning a large pretrained model on ground truth labels.
This model’s resulting performance is called the \textit{strong ceiling performance}.

To measure the fraction of the performance gap that the strong student model can recover with weak supervision, ~\citet{burns2023weak} define the performance gap recovered (PGR) using the above three performances:

\begin{equation} 
\mathrm{PGR}= \frac{\text{weak-to-strong}-\text{weak}}{\text{strong ceiling}-\text{weak}}. \nonumber
\end{equation}

To bridge gap between weak-to-strong performance and strong ceiling performance, they propose an auxiliary confidence loss that reinforces the strong model’s confidence in its own predictions:
\begin{equation} 
\small
\mathcal{L}_{\mathrm{conf}}=\mathrm{CE}(f(x), (1-\alpha)\cdot f_w(x)+\alpha \cdot \hat{f}_t(x)), \nonumber
\end{equation}
where $\mathrm{CE}(\cdot,\cdot)$ is the cross-entropy loss between two distributions on a given input $x$, $f(x)\in[0,1]$ is the strong model predictive distribution, $f_w(x)\in[0,1]$ is the weak label predictive distribution, $\hat{f}_t(x)=I[f(x)>t]\in\{0,1\}$ is the hardened strong model prediction using a threshold $t$ where $I$ is the indicator function. $\alpha$ is a (fixed) weight (usually $0.5$) to produce the cross-entropy target.

\begin{figure*}[t]
\centering
\includegraphics[width = 0.7\linewidth,height=4cm]{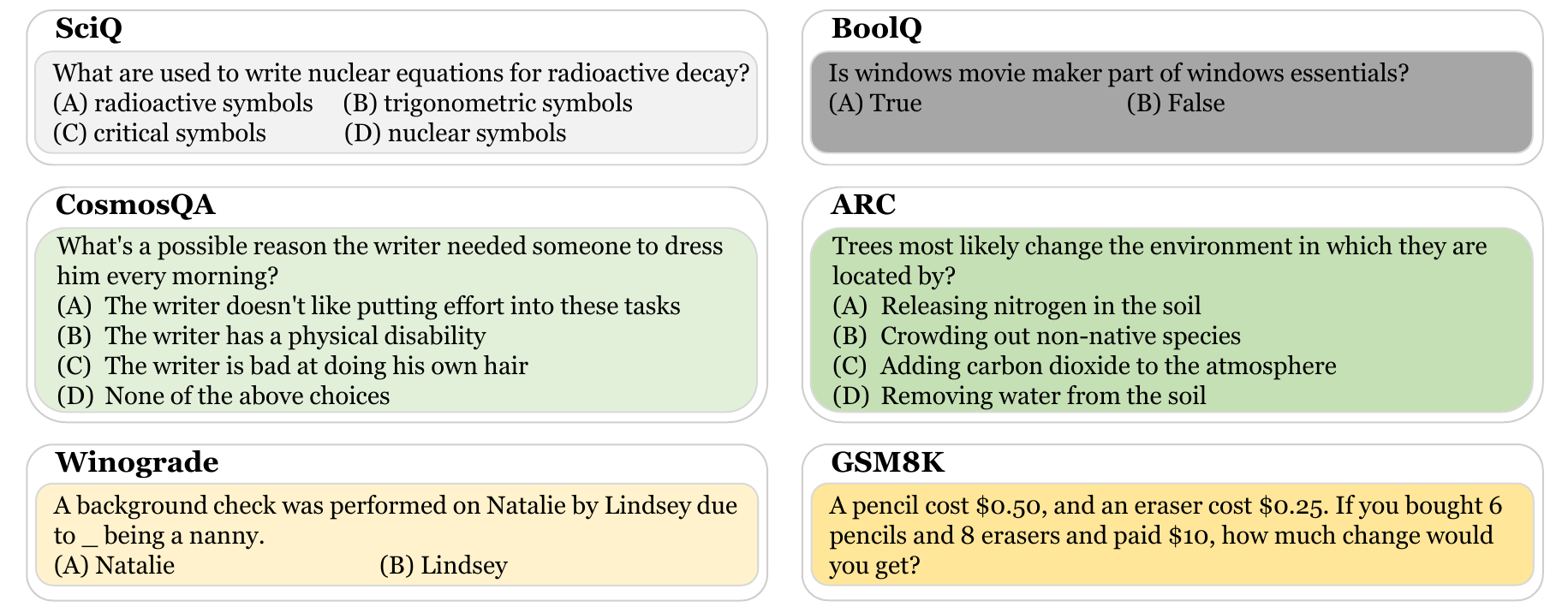}
\caption{\textbf{Example problems from diverge NLP tasks}. We present problems from SciQ, BoolQ, CosmosQA, ARC, Winograde, and GSM8K datasets.}
\label{fig:example}
\end{figure*}

\section{Methods}
\subsection{Overview}
In this study, we build the strong student model following three steps:
\textbf{1.} Train models to predict whether they can answer questions correctly;
\textbf{2.} Estimate graph-smoothed weak labels for questions that a model cannot answer;
\textbf{3.} Train a strong student model using graph-smoothed weak labels and self-generated labels selectively.

\subsection{Training to Predict P(IK)}

Central to our framework is the idea of \textit{selective weak-to-strong generalization}, where the system decides whether the strong student model could benefit from weak supervision and abstains from training with weak labels when it is deemed unnecessary (Figure~\ref{fig:diagram} bottom).

Concretely, after observing the current question, the strong model explicitly predicts whether or not it can correctly answer the question.
If the strong model knows the correct answer, we train the strong model with labels generated by itself; otherwise, we still train the strong model with weak labels.

This approach is inspired by the findings in~\citet{kadavath2022language}. 
They show that strong pretrained models can be trained to predict whether they know the answer to any given question, denoting the probability they assign as P(IK) (for Probability that I Know the answer).
This is fundamentally a question about the strong models themselves, which demonstrate their abilities in directly evaluating their own knowledge state.

To effectively train P(IK), we create a training set in the form of (few-shot prompt + question, ground truth label).
There are just two choices for the label (IK / IDK), where IK means the model knows the correct answer, and IDK means the model does not know.
For a given question $Q$, if the model's sampled answer is correct, our training set contains a datapoint ($Q$, IK); otherwise, a datapoint ($Q$, IDK).
We use a 4-shot prompt simply to ensure that the model's sampled answers are in the correct format.

We implement a binary classifier P(IK) by equipping the model with a linear classification head with two outputs.
During P(IK) training, we finetune the entire model along with the head using a cross-entropy loss $\mathcal{L}_{\mathrm{ik}}$.
Table~\ref{tab:prompt2} shows some examples of P(IK) scores from a 14 billion parameter model on a few example questions where the model sensibly should or should not know the answers.

As a note, in a later section we study out-of-distribution generalization and easy-to-hard generalization of P(IK).
We believe that generalization of P(IK) is crucial for selective weak-to-strong generalization because it will be applied in future novel and complex tasks.

\begin{table*}[t]
    \centering
    \small
    
    \begin{tabular}{l|l|cc|cc|cc}
    \toprule
    \multicolumn{1}{l|}{\multirow{2}{*}{\textbf{Performance}}} & \multicolumn{1}{l|}{\multirow{2}{*}{\textbf{Method}}} & \multicolumn{2}{c|}{\textbf{SciQ}} & \multicolumn{2}{c|}{\textbf{BoolQ}} & \multicolumn{2}{c }{\textbf{CosmosQA}} \\
    \multicolumn{1}{l|}{} & \multicolumn{1}{l|}{} &  Acc. &  PGR &  Acc. &  PGR &  Acc. &  PGR \\
    \midrule
    Weak performance & & 64.70 &   & 64.91 &  & 67.17     \\
    \midrule
    \multirow{4}{*}{Weak-to-strong performance} & Finetune & 77.90 & 45.99 & 65.89 & 3.92& 71.99 & 21.37\\
    & Finetune w/  auxi. loss & 82.20 & 60.98 & 62.22 & -10.76& 73.37 & 27.49 \\
    & Finetune w/ prod. loss & 81.60 & 58.89 & 61.02 & -15.56& 73.40 & 27.63 \\
    & Finetune w/ adap. loss & 74.00 & 32.40 & 62.83 & -8.32& 67.77 & 2.66 \\
    & Finetune w/ rkl loss & 78.50  & 48.08  & 66.29  & 5.52 & 74.10  & 30.73  \\
    & Finetune w/ js loss & 80.00   & 53.31  &  65.80 & 3.56 & 71.96  & 21.24  \\
    & \textbf{Ours} & \textbf{83.30} & \textbf{64.81} & \textbf{68.64} & \textbf{14.92} & \textbf{75.11} & \textbf{35.21}\\
    \midrule
    Strong ceiling performance & & 93.40 &   & 89.91 &  & 89.72 &     \\
    \bottomrule
    \end{tabular}
    \caption{\textbf{Selective weak-to-strong generalization improves generalization}. Test accuarcy (\%) and performance gap recovered (PGR) (\%) of our approach and baselines on the binary classification tasks converted from NLP classification datasets. Here, we use \textbf{GPT-2}~\citep{radford2019language} for the weak model and \textbf{Qwen/Qwen-14B}~\citep{bai2023qwen} for the strong model, respectively. Accuracy of weak and strong models trained with ground truth are reported as weak performance and strong ceiling performance, respectively.}
    \label{tab:main_result_w_gpt2}
\end{table*}

\begin{table*}[t]
    \centering
    \small
    
    \begin{tabular}{l|l|cc|cc|cc}
    \toprule
    \multicolumn{1}{l|}{\multirow{2}{*}{\textbf{Performance}}} & \multicolumn{1}{l|}{\multirow{2}{*}{\textbf{Method}}} & \multicolumn{2}{c|}{\textbf{SciQ}} & \multicolumn{2}{c|}{\textbf{BoolQ}} & \multicolumn{2}{c }{\textbf{CosmosQA}} \\
    \multicolumn{1}{l|}{} & \multicolumn{1}{l|}{} &  Acc. &  PGR &  Acc. &  PGR &  Acc. &  PGR \\
    \midrule
    Weak performance & & 83.80 &   & 78.92 &  & 79.13     \\
    \midrule
    \multirow{4}{*}{Weak-to-strong performance} & Finetune & 88.10 & 44.79 & 80.85 & 17.56& 81.37 & 21.15\\
    & Finetune w/  auxi. loss & 89.50 & 59.38 & 82.53 & 32.85& 81.34 & 20.87 \\
    & Finetune w/ prod. loss & 89.40 & 58.33 & 82.47 & 32.30& 82.95 & 36.07 \\
    & Finetune w/ adap. loss & 89.60 & 60.42 & 81.79 & 26.11& 80.13 & 9.44 \\
    & Finetune w/ rkl loss & 88.60  & 50.00  & 82.62  & 33.67  & 82.45  & 31.35  \\
    & Finetune w/ js loss & 88.80  & 52.08  &  83.33 & 40.13 & 81.84  & 25.59  \\
    & \textbf{Ours} & \textbf{90.60} & \textbf{70.83} & \textbf{83.39} & \textbf{40.67} & \textbf{83.85} & \textbf{44.57}\\
    \midrule
    Strong ceiling performance & & 93.40 &   & 89.91 &  & 89.72 &     \\
    \bottomrule
    \end{tabular}
    \caption{\textbf{Selective weak-to-strong generalization improves generalization}. Test accuarcy (\%) and performance gap recovered (PGR) (\%) of our approach and baselines on the binary classification tasks converted from NLP classification datasets. Here, we use \textbf{Qwen/Qwen-1.8B} for the weak model and \textbf{Qwen/Qwen-14B} for the strong model, respectively~\citep{bai2023qwen}. Accuracy of weak and strong models trained with ground truth are reported as weak performance and strong ceiling performance, respectively.}
    \label{tab:main_result_w_1_8b}
\end{table*}

\begin{table*}[t]
    \centering
    \small
    \begin{tabular}{l|cc|cc|cc}
    \toprule
    \multicolumn{1}{l|}{\multirow{2}{*}{\textbf{Method}}} & \multicolumn{2}{c|}{\textbf{SciQ}} & \multicolumn{2}{c|}{\textbf{BoolQ}} & \multicolumn{2}{c}{\textbf{CosmosQA}} \\
    \multicolumn{1}{l|}{} &  Acc. &  PGR &  Acc. &  PGR &  Acc. &  PGR  \\
    \midrule
    Ours w/o IK & 89.86 & 63.13 & 82.28 & 30.57 & 81.36 & 21.06 \\
    Ours w/o GS & 90.23 & 66.98 & 83.11 & 38.13& 83.12 & 37.68 \\
    MTL & 89.98 & 64.38 & 82.56 & 33.12 & 82.27 & 29.65 \\
    \midrule
    \textbf{Ours} & \textbf{90.60} & \textbf{70.83} & \textbf{83.39} & \textbf{40.67} & \textbf{83.85} & \textbf{44.57}\\
    \bottomrule
    \end{tabular}
   \caption{Ablation on main components of our model.}
    \label{tab:ab_compo}
\end{table*}
\subsection{Estimating Graph-smoothed Weak Labels}

To further capitalize on the ability of a strong model in predicting P(IK), we refine weak labels with a graph smoothing method~\citep{luo2018smooth,lang2022estimating}.

Given predictions of P(IK) in the current batch, questions to train the strong model can be divided into two groups: IK questions $\mathcal{D}_\mathrm{ik}$ with P(IK) scores above a threshold $\gamma$ and IDK questions $\mathcal{D}_\mathrm{idk}$ with P(IK) scores below $\gamma$.
Note that IK questions have reliable labels generated by the strong model, and IDK questions have unreliable weak labels provided by the weak supervisor.

We construct a fully connected unidirectional embedding graph $\mathcal{G}$ using samples in $\mathcal{D} = \mathcal{D}_\mathrm{ik} \cup \mathcal{D}_\mathrm{idk}$.
We first map each sample $x \in \mathcal{D}$ into an embedding $z$ (computed from the final transformer layer), and then use all these embeddings as nodes for $\mathcal{G}$.
We also assign a prior label $l_p(x)$ to each sample $x \in \mathcal{D}$ to represent its annotation, i.e., for a sample $x \in \mathcal{D}_\mathrm{ik}$, $l_p(x)$ is defined as the self generated label by the strong model, and for a sample $x \in \mathcal{D}_\mathrm{idk}$, $l_p(x)$ is defined as the corresponding weak label.

For each sample $x \in \mathcal{D}_\mathrm{idk}$, a graph-smoothed label $l_g(x)$ is obtained by aggregating adjacent nodes on $\mathcal{G}$.
Specifically, to conform to the smoothness assumption, we try to minimize the following distance when determining $l_g(x)$:
\begin{equation}
\small
\begin{split}
& \alpha \cdot d[l_g(x), l_p(x)] + (1-\alpha) \sum_{x_j \in \mathcal{D}} a_{j} \cdot d[ l_g(x), l_p(x_j) ] \label{eq:lable_obj} \\
& a_{j} = \frac{{\rm exp}(z \cdot z_j / \tau)}{\sum_{k=1}^{|\mathcal{D}|} {\rm exp}(z \cdot z_k / \tau)},
\end{split}
\end{equation}
where $0 \leq \alpha \leq 1$ is a weight, $d$ is a distance function, $\tau > 0$ is a temperature. The second term in Eq. \ref{eq:lable_obj} enforces the smoothness assumption by encouraging $l_g(x)$ to have similar labels with its nearby samples, whereas the first term tries to maintain $l_g(x)$ to meet its original annotation $l_p(x)$. For simplicity, we implement $d$ as the Euclidean distance here, and thus minimizing Eq. \ref{eq:lable_obj} yields:
\begin{equation}\label{eq:graph_smooth}
\small
l_g(x) = \alpha \cdot l_p(x) + (1-\alpha) \sum_{x_j \in \mathcal{D}} a_{j} \cdot l_p(x_j)
\end{equation}

Note that the result we derived in Eq. \ref{eq:graph_smooth} follows most previous graph-smoothing methods in semi-supervised learning \cite{van2020survey}.

\subsection{Training Models using Labels Selectively}
After producing the graph-smoothed weak labels, we construct a dataset with improved labels:
\begin{equation}
\small
\label{eq-m}
\mathcal{M}=\{(x_i, l_p(x_i)) \cup (x_j, l_g(x_j))|x_i \in \mathcal{D}_\mathrm{ik}, x_j \in \mathcal{D}_\mathrm{idk} \}
\end{equation}

We can finetune a model on $\mathcal{M}$ with a cross-entropy loss $\mathcal{L}_{\mathrm{gen}}$.
The final training loss for finetuning a large pretrained student model is:
\begin{equation}
\small
\mathcal{L}=\mathcal{L}_{\mathrm{gen}} + \lambda \cdot \mathcal{L}_{\mathrm{ik}},
\label{eq-a}
\end{equation}
where $\mathcal{L}_{\mathrm{gen}}$ is optimized on a dataset to evaluate weak-to-strong performance, and $\mathcal{L}_{\mathrm{ik}}$ is optimized on another dataset to train P(IK), $\lambda$ is a weight.

\begin{table}[h]
    \centering
    \small
    \begin{tabular}{l|c|c}
    \toprule 
    \textbf{Method} & \textbf{SciQ} & \textbf{BoolQ} \\
    \midrule
    Finetune & 88.10 & 80.85 \\
    \midrule
    \textbf{Ours} (+CosmosQA) & \textbf{90.70} & 83.20 \\
    \textbf{Ours} (+Winograde) & 89.80 & 82.99 \\
    \textbf{Ours} (+ARC)  & 90.60 & \textbf{83.39} \\
    \bottomrule
    \end{tabular}
   \caption{\textbf{Ablation on the dataset to train the P(IK) classifier}. The performance is evaluated with accuracy.}
    \label{tab:ab_classifier}
\end{table}

\section{Experiments}

\subsection{Tasks}

We adopt the evaluation protocol of prior work~\cite{burns2023weak}, and conduct experiments in NLP tasks on three benchmark datasets: SciQ~\cite{welbl2017crowdsourcing}, BoolQ~\cite{clark2019boolq}, and CosmosQA~\cite{huang2019cosmos}.
We convert each dataset to a binary classification problem.
For multiple-choice datasets, given a data point with a question $Q$ and $k$ candidate answers $A$, we construct $k$ new data points of the form $(Q, A_i)$, where the label is $1$ for the correct
answers and $0$ for all the incorrect answers.
We also keep the same number of correct and incorrect answers per question to maintain class balance.
Figure~\ref{fig:example} shows example problems from the above-mentioned datasets.

\subsection{Experimental Setups and Metrics}

Following~\cite{burns2023weak}, we randomly sample at most 20k data points from each task and split them in half.
We train a weak model on the first half of the data points and use its prediction on the other half as the weak labels.
The weak labels are soft labels~\cite{hinton2015distilling}.
We report the accuracy and performance gap recovered (PGR) of the strong student model on the test set in all tasks.

\begin{table*}[t]
    \centering
    \small
    
    \begin{tabular}{l|cc|cc}
    \toprule
    \multicolumn{1}{l|}{\multirow{2}{*}{\textbf{}}}  & \multicolumn{2}{c|}{\textbf{In-Dist Generalization}} & \multicolumn{2}{c}{\textbf{OOD Generalization}}  \\
    \multicolumn{1}{l|}{}  &  Training Data &  AUROC &  Training Data &  AUROC  \\
    \midrule
    SciQ & SciQ & 92.14  & ARC & 89.34       \\
    BoolQ & BoolQ & 85.08 & ARC & 79.02 \\
    CosmosQA & CosmosQA & 88.01 & ARC & 79.92 \\
    Winograde & Winograde & 73.43 & ARC & 63.55 \\
    GSM8K & GSM8K &69.27 & ARC & 47.73\\
    \bottomrule
    \end{tabular}
    \caption{\textbf{Overview comparing out-of-distribution generalization to in-distribution generalization performance of P(IK)}. All AUROC scores (\%) are computed using the Qwen/Qwen-14B P(IK) classifiers. Even when we only train on ARC, we see decent generalization to other tasks.}
    \label{tab:ik_g}
\end{table*}

\subsection{Implementation Details}

Our implementations of data preprocessing, weak and strong model training are based on the OpenAI weak-to-strong codebase and its default hyper-parameters~\cite{burns2023weak}.
Specifically, we use Qwen/Qwen-14B~\cite{bai2023qwen} as the large pretrained model for training strong models.
Meanwhile, we use GPT-2~\citep{radford2019language} and Qwen/Qwen-1.8B as two small pretrained models for training weak models, which have different gaps in compute between weak and strong models.

In order to adapt weak and strong models to the converted binary classification setting, we equip each model with another linear classification head with two outputs.
We use $\lambda=1$, $\gamma=0.8$, $\alpha=0.9$, and $\tau=0.1$ in all experiments.
We train all models for two epochs with a batch size of 32.
We conduct all experiments on a single 8×A100 machine.

\subsection{Baselines}

We compare our approach with competitive baseline approaches:
1. \textbf{Finetune}~\cite{burns2023weak} naively finetunes strong pretrained models on labels generated by a weak model;
2. \textbf{Finetune w/ auxi. loss}~\cite{burns2023weak} finetunes strong models with an auxiliary confidence loss, which reinforces the strong model’s confidence in its own predictions;
3. \textbf{Finetune w/ prod. loss}~\cite{burns2023weak} finetunes strong models with a confidence-like loss which sets the cross entropy targets to the product of weak labels and strong model predictions.
4. \textbf{Finetune w/ adap. loss}~\cite{guo2024vision} finetunes strong models with an adaptively adjustable loss using the discrepancy between the soft label and the hard label.
5. \textbf{Finetune w/ rkl loss}~\cite{yao2025revisiting} finetunes strong models with reverse KL divergence loss.
6. \textbf{Finetune w/ js loss}~\cite{yao2025weak} finetunes strong models with Jensen–Shannon divergence loss.
We also report the \textbf{weak performance} and the \textbf{strong ceiling performance} defined in the preliminaries section.
Note that the strong ceiling performance is regarded as the upper bound of the \textbf{weak-to-strong performance} when only weak labels are considered.

\subsection{Main Results}

\begin{figure*}[!h]
\centering
\includegraphics[width = 0.7\linewidth,height=14cm]{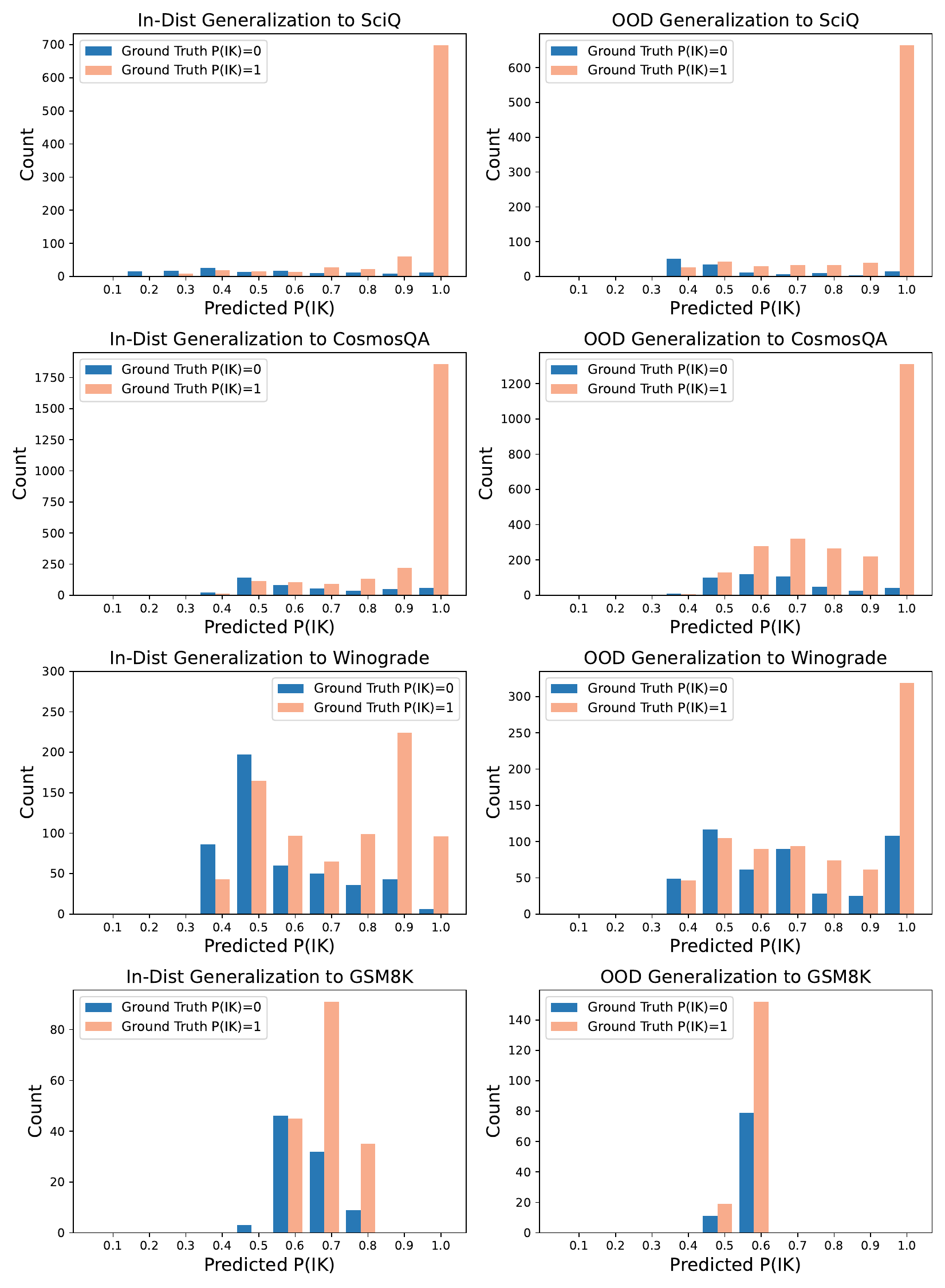}
\caption{\textbf{Generalization of P(IK)}. The left side of this figure includes distributions of P(IK) from a Qwen/Qwen-14B P(IK) classifier that was trained on in-distribution data, i.e., SciQ, CosmosQA, Winograde, and GSM8K, respectively. The right side includes distributions of P(IK) that was trained on just ARC.}
\label{fig:gen}
\end{figure*}

Table~\ref{tab:main_result_w_gpt2} and Table~\ref{tab:main_result_w_1_8b} show the results of each approach on the binary classification tasks converted from SciQ, BoolQ, and CosmosQA datasets.
Here, our approach trains the binary classier P(IK) on questions from the ARC dataset~\citep{clark2018think}.
We use GPT-2 for the weak model in Table~\ref{tab:main_result_w_gpt2} and use Qwen/Qwen-1.8B for the weak model in Table~\ref{tab:main_result_w_1_8b}.
In each task, we observe that PGRs of strong student models naively finetuned on weak labels are all positive, which indicates promising weak-to-strong generalization.

At the same time, we find that our approach significantly outperforms each strong student baseline, including the naive baseline finetuned on weak labels or more sophisticated baselines equipped with a confidence loss term on all three tasks.
Compared with the promising baseline Finetune w/ auxi. loss, our approach brings up from a PGR of $60.98\%$ to $64.81\%$ in SciQ, $-10.76\%$ to $14.92 \%$ in BoolQ, and $27.49\%$ to $35.21\%$ in CosmosQA, when using weak model GPT-2, and brings up from a PGR of $59.38\%$ to $70.83\%$ in SciQ, $32.85\%$ to $40.67 \%$ in BoolQ, and $20.87\%$ to $44.57\%$ in CosmosQA, when using weak model Qwen/Qwen-1.8B.
Our approach also obtains the best test accuracy among all compared strong students.
The performance gain shows the advantage of selective W2SG, which helps elicit the capabilities of the strong model with weak supervision.

\subsection{Ablation Studies}

We provide comprehensive ablation studies to understand the efficacy of selective weak-to-strong generalization framework.
All ablations are conducted using Qwen/Qwen-1.8B for the weak model and Qwen/Qwen-14B for the strong model.

\paragraph{Ablation on model main components.} We verify the effect of each component in our selective framework by testing following variants:
1. \textbf{Ours w/o IK} removes training the P(IK) classier. In this variant, IDK questions $\mathcal{D}_\mathrm{idk}$ contains all the questions to train the student model, and the loss shown in Eq.~\ref{eq-a} is optimized by setting $\lambda=0$;
2. \textbf{Ours w/o GS} removes estimating the graph-smoothed weak labels. In this variant, $l_g(x)$ of each sample $x\in\mathcal{D}_\mathrm{idk}$ shown in Eq.~\ref{eq-m} is replaced with the prior label $l_p(x)$.
3. \textbf{MTL} implements multi-task learning with one head for P(IK) and another head for alignment. The head for P(IK) is trained with ground truth labels from the ARC dataset while the other head is trained with weak labels.

Results in Table~\ref{tab:ab_compo} indicate that  our method outperforms all ablation models in terms of test accuracy and PGR.
We can further observe that:
\textbf{1.} Training the P(IK) classier to identify IK questions brings the largest improvement compared to other components.
This proves the importance of understanding whether a strong model knows the correct answer to a question for selective W2SG.
\textbf{2.} Naively training the P(IK) classier without selecting labels for alignment degenerates the model performance by a large margin.  
This shows the effectiveness of improved label quality produced by our selective framework.

\paragraph{Ablation on the dataset to train the P(IK) classifier.} We analyze the impact of the dataset to train the P(IK) classifier on the final performance.
In Table~\ref{tab:ab_classifier}, we train the P(IK) classifier on questions from three different datasets: CosmosQA, Winograde~\cite{sakaguchi2021winogrande}, and ARC.
We can observe that our selective W2SG method consistently outperforms methods that always use weak supervision, when training the P(IK) classifier with different extra datasets.
It also suggests that strong models can be elicited to evaluate their own knowledge state.
We present analyses of the P(IK) generalization ability in the next subsection.

\subsection{Further Analysis}\label{sec:ana}

Achieving good generalization of P(IK) is important for selective W2SG, since it will be applied in future novel and complex tasks for superalignment. 
We are interested in studying the generalization of P(IK) across various dimensions, i.e., cross-task, cross-domain, and cross-difficulty.

Specifically, we train P(IK) classifiers only on ARC and then evaluating on SciQ, BoolQ, CosmosQA, Winograde, and GSM8K~\cite{cobbe2021training}.
These datasets cover diverse domains such as science QA, reading comprehension, commonsense reasoning, and math  (see Figure~\ref{fig:example} for an illustration).
Meanwhile, these datasets also cover diverse difficulties, where Winograde and GSM8K are much harder than the other datasets, with relatively lower test accuracy (see Appendix).
Hence, evaluation on these datasets help us explore out-of-distribution generalization and easy-to-hard generalization~\citep{sun2024easy} of P(IK).
In this analysis, we implement a variant of our model that only optimizes the loss $\mathcal{L}_{\mathrm{ik}}$ in Eq.~\ref{eq-a}, i.e., $\mathcal{L}_{\mathrm{gen}}$ is removed.

Table~\ref{tab:ik_g} gives an overview of generalization performance for the P(IK) classifier that is trained on ARC, compared to its in-distribution performance.
Figure~\ref{fig:gen} gives a detailed view of how the distribution of P(IK) changes depending on training data.
We find that strong models do exhibit a degree of generalization of P(IK) across tasks and difficulties.

\section{Limitations and Conclusion}

\paragraph{Limitations.}
Although our proposed method is found to be effective in all our experiments, the difference between strong and weak models is only in the size of pretrained models in our setup.
However, in the future, strong models may also differ in reasoning and planning abilities.
Furthermore, since weak-to-strong deception phenomenon exists~\cite{yang2024super}, the P(IK) classifier might not be perfectly robust for future superalignment. It highlights the urgent need to pay more attention to alignment reliability.


\paragraph{Conclusion.}
In this paper, we challenge the assumption of always using weak supervision for W2SG.
In response, we propose a selective framework, where strong models identify whether using weak labels is necessary.
Extensive evaluations show that our approach outperforms SOTA baselines.

\bibliography{aaai2026}

\appendix

\section{Evaluation on diverse NLP tasks} 

\begin{figure}[h]
\centering
\includegraphics[width = 0.8\linewidth]{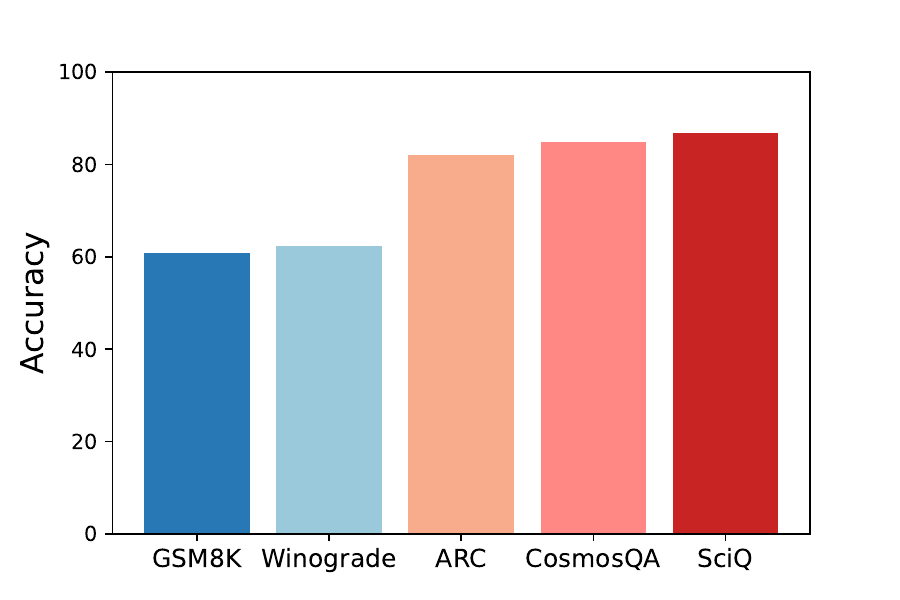}
\caption{Few-shot performances of Qwen/Qwen-14B on diverse NLP tasks.}
\label{fig:few}
\end{figure}

We evaluate few-shot performances of Qwen/Qwen-14B~\citep{bai2023qwen} on diverse NLP tasks.
Figure~\ref{fig:few} shows that Winograde and GSM8K are much harder than the other datasets, i.e., ARC, CosmosQA, and SciQ, with relatively lower test accuracy.

\end{document}